\documentclass[10pt,twocolumn,letterpaper]{article}

\usepackage[pagenumbers]{cvpr} 

\usepackage{graphicx}
\usepackage{amsmath}
\usepackage{amssymb}
\usepackage{booktabs}
\usepackage{breqn}
\usepackage{multirow}
\usepackage{tabularx}
\usepackage{array}

\usepackage[pagebackref,breaklinks,colorlinks]{hyperref}

\usepackage[capitalize]{cleveref}
\crefname{section}{Sec.}{Secs.}
\Crefname{section}{Section}{Sections}
\Crefname{table}{Table}{Tables}
\crefname{table}{Tab.}{Tabs.}

\def\cvprPaperID{11585} 
\def\confName{CVPR}
\def\confYear{2023}

\graphicspath{{./figure/}}

\begin{document}

\title{Consistency of Implicit and Explicit Features Matters for Monocular 3D Object Detection}

\author{Qian Ye, Ling Jiang, Wang Zhen, Yuyang Du \\
DiDi Chuxing\\
Beijing, China\\
{\tt\small \{noahye, jiangling, wangzhenwangzhen, duyuyang\}@didiglobal.com}
}
\maketitle

\begin{abstract}
  Low-cost autonomous agents including autonomous driving vehicles chiefly adopt monocular 3D object detection to perceive surrounding environment. This paper studies 3D intermediate representation methods~\cite{reading2021categorical} which generate intermediate 3D features for subsequent tasks. For example, the 3D features can be taken as input for not only detection, but also end-to-end prediction and/or planning that require a bird's-eye-view feature representation. In the study, we found that in generating 3D representation previous methods do not maintain the consistency between the objects' implicit poses in the latent space, especially orientations, and the explicitly observed poses in the Euclidean space, which can substantially hurt model performance. To tackle this problem, we present a novel monocular detection method, the first one being aware of the poses to purposefully guarantee that they are consistent between the implicit and explicit features. Additionally, we introduce a local ray attention mechanism to efficiently transform image features to voxels at accurate 3D locations. Thirdly, we propose a handcrafted Gaussian positional encoding function, which outperforms the sinusoidal encoding function~\cite{NIPS2017_3f5ee243} while retaining the benefit of being continuous. Results show that our method improves the state-of-the-art 3D intermediate representation method by 3.15\%. We are ranked 1st among all the reported monocular methods on both 3D and BEV detection benchmark on KITTI leaderboard as of this paper's submission time. 
\end{abstract}

\begin{figure*}
  \centering
  \includegraphics[width=\textwidth]{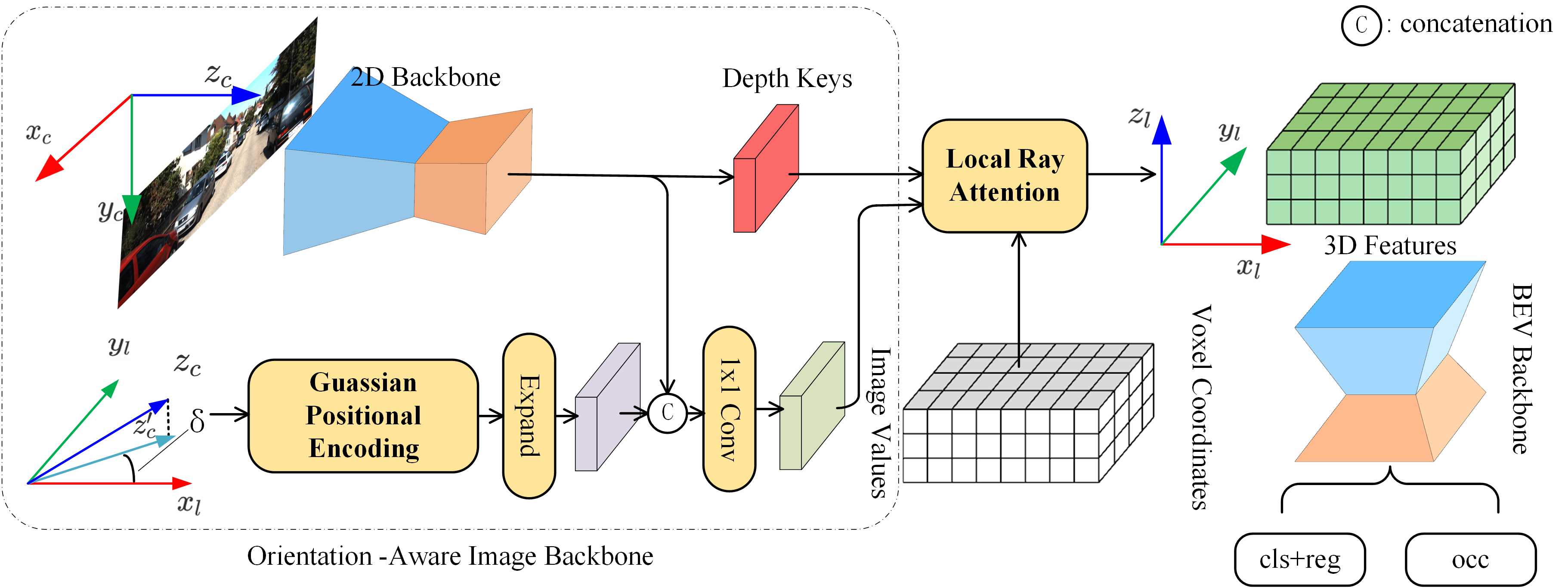}
  \caption{Overall Architecture of our model. The orientation-aware image backbone produces depth keys and image values from a single image and $\delta$(direction of $z_c'$ in $x_l$-$y_l$ plane). $z_c'$ is the projection of $z_c$ axis onto the $x_l$-$y_l$ plane, where subscript $l$ and $c$ indicate ego-car and camera coordinate system respectively. The local ray attention mechanism takes as input voxel coordinates, depth keys, and image values to generate a 3D representation. The 3D representation is then used by the BEV Backbone to predict 3D bounding boxes and a 3D occupancy map.}
  \label{fig:overall_architecture}
\end{figure*}

\section{Introduction}
\label{sec:intro}
Monocular 3D object detection is widely used in autonomous driving systems and in particular advanced driver-assistance systems (ADAS) to perceive the external environment. In such a setting, images captured by one camera are the only sensory input to estimate an object's location, size, and orientation. Despite the fact that most vehicles are equipped with multiple cameras, the capability to sense through monocular 3D object detection remains indispensable, as a large fraction of the surroundings is visible to only one camera. Detecting objects precisely in 3D space, such as cars, pedestrians, cyclists etc., is the basic requirement for vehicles to perform subsequent driving tasks, such as collision avoidance and adaptive cruise control.

Monocular 3D detection often utilizes the 3D intermediate representation method\cite{10.1007/978-3-030-58568-6_12, BMVC2019, reading2021categorical, weng2019monocular, ma2019accurate}, which gathers projected image features in 3D space for the first stage, and then the 3D features are consumed by the second detection stage to estimate 3D bounding boxes. This intermediate 3D scene representation has inherent advantages\cite{10.1007/978-3-030-58568-6_12, BMVC2019, reading2021categorical}: (a) Euclidean distances, such as how far away a pedestrian is, are more representative in 3D coordinate system than in 2D, (b) since scales of the objects are depth-independent, they can be better detected in 3D space [23, 26, 27]. Additionally, a detection pipeline can further process the 3D representation to a bird's-eye-view feature map which is increasingly required as input to SOTA end-to-end prediction and planning paradigms.

For existing 3D representation methods, camera parameters are used in conjunction with pixel depth estimates from image-based networks to transform features from a 2D image plane to a 3D space as points or voxels. However, by directly polulate points/voxels with RGB values or raw feature vectors from the canonical 2D image backbone, they all neglect or limit the power of the 2D image backbone. In the first case where RGB values are fed into 3D space, the image backbone only serves the purpose of a depth estimator, which other than pixels' depths and RGB values does not contribute anything to 3D representation. In the second case, each point or voxel has a raw feature vector that contains implicit features in latent space, but the implicit features can be inconsistent with the explicit features which are extracted by comparing points or voxels in Euclidean space. For example, the orientation of an object observed in an image, which is subsequently transformed to be part of an implicit feature vector by the image backbone, might conflict with the orientation macroscopically observed in Euclidean space. This is because mounting errors of cameras and the need for 3D world rotation augmentation \cite{10.1007/978-3-030-58568-6_12, yan2018second} will cause large orientation variance as a result between camera coordinate and the ego-car coordinate. Thus, the following 3D network (\eg, 3D or bird's-eye-view convolution, PointNet~\cite{qi2017pointnet}) will receive conflict information. 

We argue that there should be consistency between the implicit features in latent space and the explicit features in Euclidean space, where the latent space is composed of all possible vectors of a particular point or voxel. For the detection task, if there are conflicting features in location and especially in orientation of an object, the model's performance can be substantially impacted. Therefore, after studying the influence of orientation inconsistency, we propose a novel end-to-end orientation-aware detection method that bridges the gap between 2D and 3D representations. 

The main contributions of our work are as follows:

\begin{enumerate}
\setlength\itemsep{0em}
\item We present an orientation-aware image backbone to eliminate the disparity between implicit and explicit orientations of objects in 3D representation. 
\item We introduce a local ray attention mechanism that efficiently and directly transforms image-view features to voxel representations.
\item We propose a handcrafted Gaussian positional encoding function that outperforms the sinusoidal positional encoding function but maintain the benefit of being continuous.
\end{enumerate}

Experiments in the KITTI object detection benchmark show that our model achieves top results in two challenging tasks (ranked first on the BEV benchmark and first on the 3D benchmark). Our model outperforms the second-best method in all metrics, especially with a gain of 6.56\% in $AP_{BEV}$ {\it Easy}.

\begin{figure*}
  \centering
  \includegraphics[width=\textwidth]{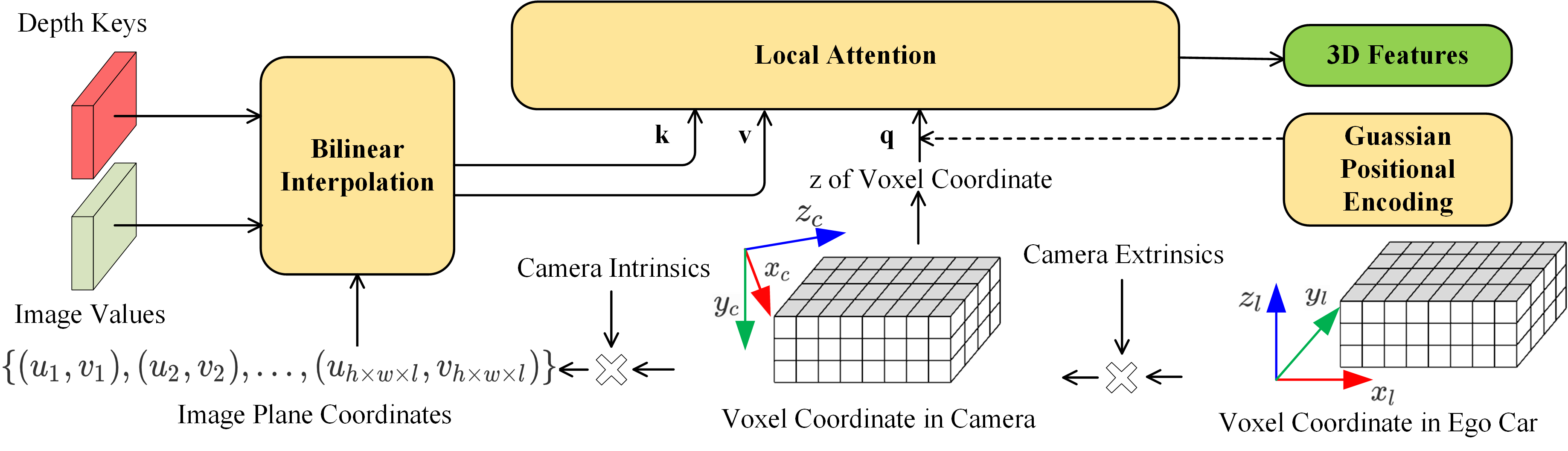}
  \caption{An illustration of the local ray attention mechanism. $(u, v)$ indicates the pixel coordinates of the projected voxel center.}
  \label{fig:local_ray_attention}
\end{figure*}

\section{Related Work}
\label{sec:related_work}
\subsection{Direct Methods}
Traditional monocular detectors use only frontal-view images as input to directly infer 3D bounding boxes from grids in the image plane without any extra sensory inputs. These methods try to relate the object geometry of the 2D image plane to the 3D space with only cues from object shapes and scene geometry. SMOKE~\cite{liu2020SMOKE} modifies a 2D detection network and proposes a simple architecture for 3D detection built on CenterNet~\cite{zhou2019objects}. FCOS3D~\cite{wang2021fcos3d} extends the 2D detector FCOS~\cite{tian2019fcos} by adding heads to predict various 3D cuboid parameterizations. MonoCon~\cite{Liu2021LearningAM} adds auxiliary context regression heads to assist 3D detection. DETR3D~\cite{detr3d}, based on DETR~\cite{carion2020end}, employs a transformer for end-to-end 3D bounding box prediction by projecting learnable 3D queries in 2D images. MonoDTR~\cite{Huang2022MonoDTRM3} learns depth information by a depth-aware transformer. To further improve detection accuracy in terms of geometry, some methods have introduced geometric priors into the networks. MonoPair~\cite{monopair} leverages geometric consistency between adjacent object pairs, formulating the inference as a constrained optimization problem. Nevertheless, the preconceived assumption about the prior may fail in many practical scenarios. MonoDDE~\cite{Li2022DiversityMF} proposes a depth-solving system that produces 20 diverse depths in different ways to alleviate the problem of depth assumption failure. These methods only produce sparse depths for every target learning from ground truth anchors whose labeling process contains non-negligible noise and is time-consuming. In order to effectively leverage large-scale annotation-free point cloud, DD3D~\cite{Park2021IsPN} adds a depth prediction head to generate depth values for every pixel in an image during training. However, due to the perspective principle, objects that vary greatly in size may appear to have the same size in an image, which substantially hurts the accuracy.

\subsection{3D Intermediate Representation Methods}
Representing features in 3D space, such as 3D voxels, bird's-eye-view grids, or pseudo-LiDAR points, overcome the limitations of perspective range view, \eg occlusion or variability in the scale of the same object. OFT~\cite{BMVC2019} introduces orthographic feature transform to project a fixed cube of voxels onto image features to generate the 3D-voxel representation of the scene. BEVFormer~\cite{li2022bevformer} learns bird's-eye-view representation from multi-camera images via spatiotemporal transformers to capture information across cameras. Image features can also be converted to 3D pointclouds, known as the Pseudo-LiDAR method, by leveraging depth information from various depth estimators and camera parameters. Mono3DPLiDAR~\cite{weng2019monocular} applies a bounding box constraint in terms of consistency to mitigate the local misalignment and long tail issues caused by noisy pseudo-LiDAR. Ma \etal~\cite{ma2019accurate} transforms the 2D image to pseudo pointclouds and embeds the complementary RGB cue into the generated pointclouds. However, these methods rely heavily on the quality of the image-to-LiDAR generation, which suffer from their sensitivity to depth accuracy. Lift-Splat~\cite{10.1007/978-3-030-58568-6_12} learns depth distributions in an unsupervised manner to generate the pseudo-LiDAR points. Based on Lift-Splat, BEVDet~\cite{Huang2021BEVDetHM} applies an instance-level data augmentation to prevent overfitting. CaDDN~\cite{reading2021categorical} predicts categorical depth distributions for each pixel to construct the pseudo pointclouds and applies a LiDAR-based 3D detector to predict 3D bounding boxes. Nevertheless, the conversion of the pseudo-LiDAR points into 3D voxels can lead to the loss of information in these methods.

\section{Method}
\label{sec:method}
\subsection{Overview}
Our work is composed of an orientation-aware image backbone, a local ray attention, and a BEV backbone, as illustrated in \Cref{fig:overall_architecture}.

Given a single image captured by a color camera mounted on top of a vehicle, the 2D backbone generates two feature maps, namely depth keys and image features. Gaussian positional encoding function is applied to encode the implicit and explicit orientation difference as a vector, which is subsequently expanded to the same size as the image features. The image values are produced by concatenating the image features with the expanded orientation difference just mentioned. The depth keys, image values, and voxel coordinates are fed to local ray attention in order to produce 3D representation of the scene. Based on the 3D representation, the BEV backbone predicts 3D bounding boxes and a 3D occupancy map.

\subsection{Local Ray Attention}
The local ray attention directly transforms feature map of an image to 3D voxel features without intermediate pseudo pointcloud (see \Cref{fig:local_ray_attention}). By taking advantage of the attention mechanism, this process is capable of placing 2D features at appropriate locations in 3D space.

The inputs to the local ray attention are depth keys and image values from the orientation-aware backbone and the center coordinates of voxels obtained by evenly dividing a predefined detection area. Through timing camera extrinsic, the voxel center coordinates are transformed from the ego-car's coordinate system to the camera's coordinate system. Given camera intrinsics, we further get the pixel coordinates of every voxel center by projecting them to the image plane. Bilinearly interpolating depth keys and image features at the projected pixel coordinates respectively produces keys and values respectively for local attention. The $z$ coordinate of the voxel center in the camera's coordinate system are embedded by the Gaussian positional encoding function (introduced in \cref{subsec:gaussian_positional_encoding}) as a query.

Given keys, values, and queries, local attention firstly projects the keys and values back to their corresponding voxels. Then every voxel is populated with a value weighted by the similarity between its query and the key.

\begin{equation}
  \vec{\hat{v}} = sim(\vec{k}, \vec{q})\vec{v}
\end{equation}

\subsection{Orientation-Aware 2D Image Backbone and Augmentation}
\label{subsec:orientation_aware}
\begin{figure}
  \centering
  \begin{subfigure}{8cm}
    \includegraphics[width=8cm]{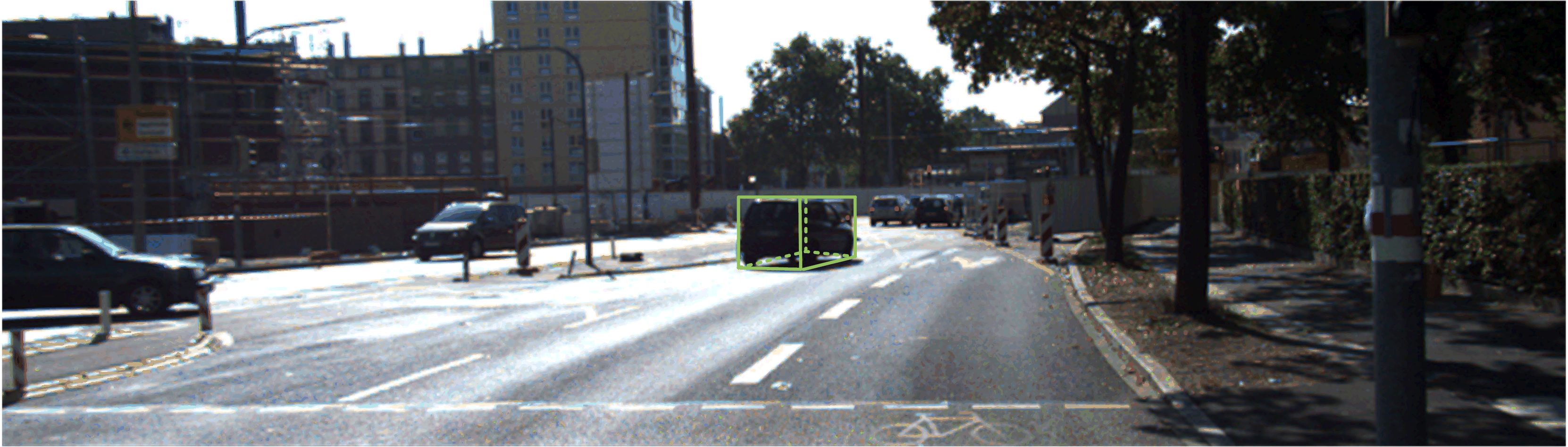}
    \caption{An image captured.}
    \label{fig:orientation_aware_raw_image}
  \end{subfigure}
  \hspace{0.5cm}
  \begin{subfigure}{8cm}
    \includegraphics[width=8cm]{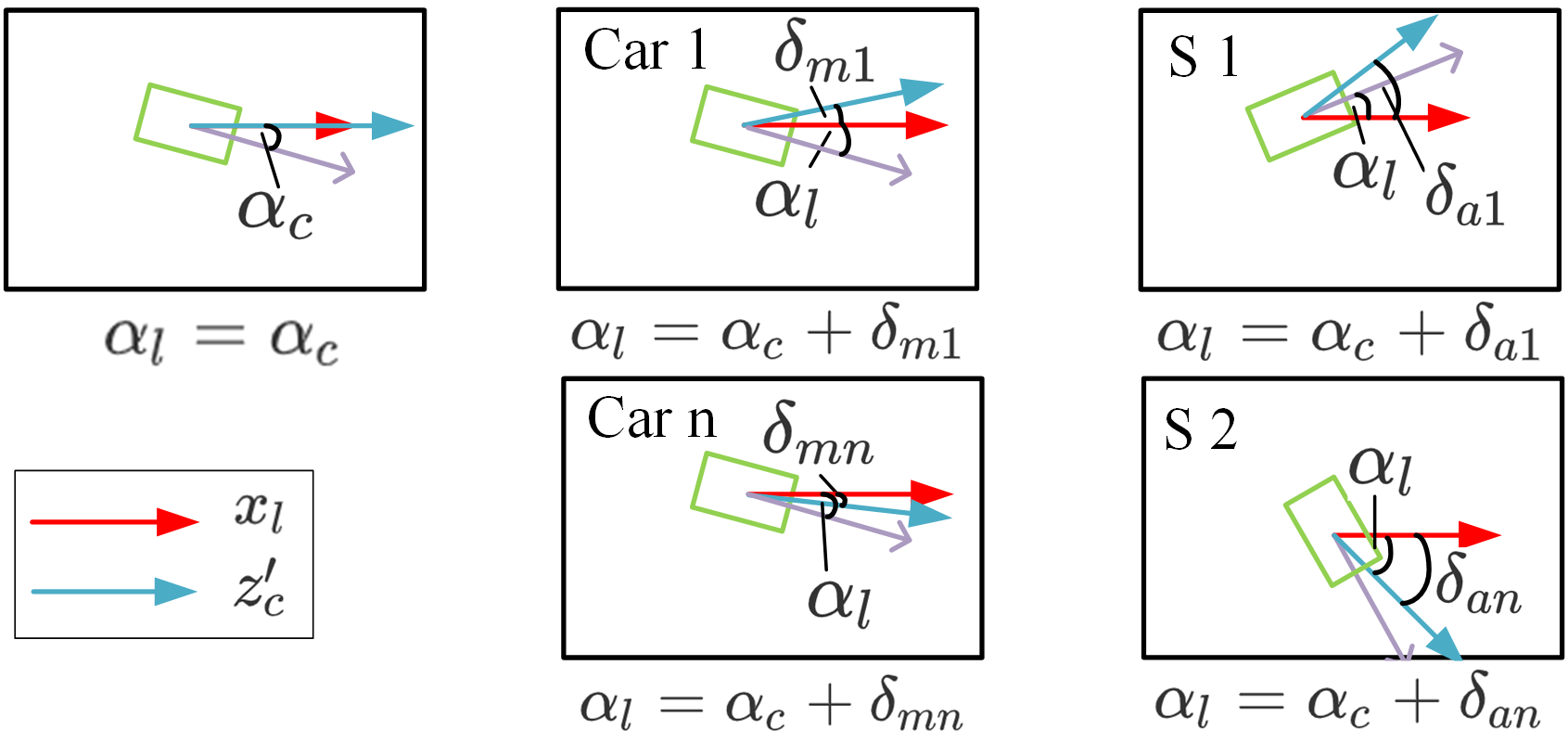}
    \caption{Different cases from a bird's-eye-view perspective, where $\alpha_l$ is the orientation of an object observed explicitly in ego-car coordinate, $\alpha_c$ is the orientation observed from a camera, and $\delta$ is the orientation of $z_c'$ in $x_l-y_l$ plane.}
    \label{fig:orientation_aware_cases}
  \end{subfigure}
  \caption{Example of implicit versus explicit orientations.}
  \label{fig:orientation_aware}
\end{figure}

The orientation-aware backbone is necessary in two cases where differences between the implicit and explicit orientations are caused as shown in \Cref{fig:orientation_aware}. 

\noindent {\bf Mounting Errors.} Ideally, in 3D intermediate representation, the implicit orientation of an object embedded by an image backbone is consistent with the explicit orientation observed in Euclidean space in ego-car's coordinate system, when $z_c'$ is aligned with $x_l$ \footnote{Since the KITTI dataset we use considers only yaw, we refer to the orientation of an object only as that on the $x_l$-$y_l$ plane.}(the first column of \cref{fig:orientation_aware_cases}). However, the detection model should handle the data collected by different ego cars in either training or deployment, resulting in different disparities between implicit and explicit orientations due to mounting errors of cameras (the second column of \cref{fig:orientation_aware_cases}). For instance, while the explicit orientation of the object remains unchanged for different ego cars in the same scene with the same pose, the orientations observed by cameras may differ depending on the direction of the camera. 

\noindent {\bf Augmentation.} During the training process, to prevent the BEV backbone from overfitting, data augmentation (i.e. random world rotation) is applied to the 3D representation. It leads to a large deviation in explicit orientation from implicit orientation (the third column of \cref{fig:orientation_aware_cases}), which cannot be corrected by image rectification because the image will be greatly compressed on a certain axis when a large random world rotation is applied.

To solve the inconsistency problem, we propose an orientation-aware image backbone to tackle the orientation inconsistency caused by the mounting errors and augmentation at the same time. The orientation-aware image backbone takes as input not only an image, but also the orientation differences to generate image values. In our settings of data augmentation (Details are in the last paragraph of this section.), the orientation difference of an object can be calculated from camera extrinsic matrix. After data augmentation, we get the ego car coordinate system, with which axes of 3D voxels are aligned. Suppose $ z_c' $ is the projection of $ z_c $ axis onto the $x_l - y_l $ plane, we calculate the orientation difference $ \delta $ of $ z_c' $ and $ x_l $ with the camera extrinsics. 

\begin{equation}
  \delta = arctan(\frac{t_{3, 2}}{t_{3, 1}})
\end{equation}
,where t is the element of camera extrinsic matrix $\mathbf{T}$ that transforms from the ego-car to the camera coordinate system.

$ \delta $ is then encoded by handcrafted Gaussian positional encoding(introduced later in \cref{subsec:gaussian_positional_encoding}) to get the vector representation, $p_\delta$. Note that We do not perform instance augmentation \cite{yan2018second, Huang2021BEVDetHM} so that the orientation differences are the same for all locations. Therefore, $ p_\delta $ is simply expanded to the same size as the image feature map (produced by the 2D backbone) and then is concatenated with it. Following a layer of 1x1 convolution, the image values containing the implicit features that are consistent with the explicit features are obtained.

Data augmentation will affect the consistency as aforementioned, so we discuss our deployed augmentation strategies here. We use random world flipping and random world rotation augmentation. For the random world flipping, ground truth bounding boxes, and the lidar points flip around the x-axis of ego-car coordinate system. The second column of $\mathbf{T}$ is set to the opposite. Image is flipped horizontally with the intrinsic matrix modified accordingly. For the random world rotation, we rotate the ground truth bounding boxes and lidar points with rotation matrix $R$. The camera extrinisc matrix is modified by multiplying the inverse of $R$ as the following equation.

\begin{equation}
  \hat{\mathbf{T}} = \mathbf{TR^{-1}}
\end{equation}

\begin{figure}
  \centering
  \begin{subfigure}{3.7cm}
    \includegraphics[width=3.7cm]{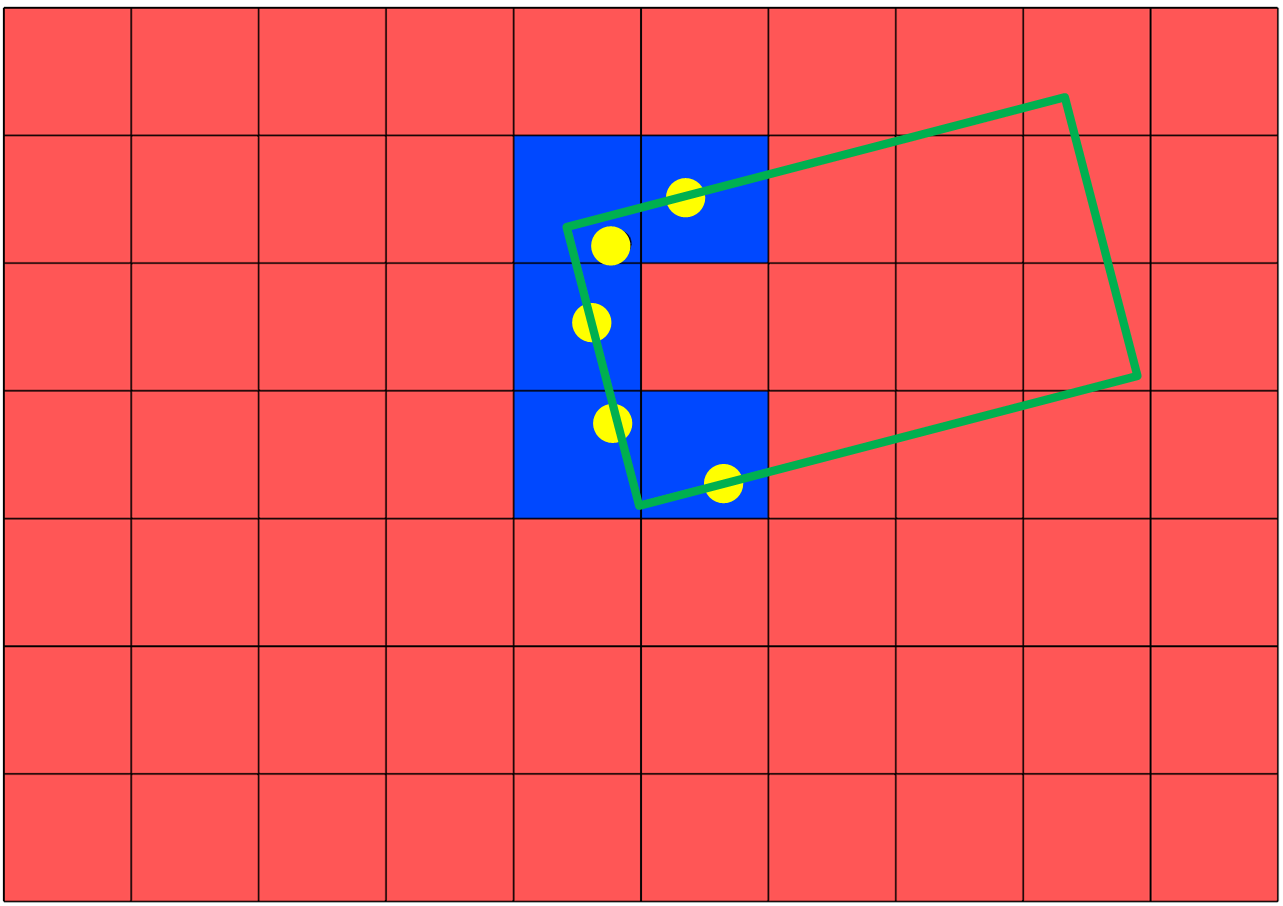}
    \caption{Taking voxels without any points as empty spaces results in false negatives.}
    \label{fig:occ_fn}
  \end{subfigure}
  \hspace{0.5cm}
  \begin{subfigure}{3.7cm}
    \includegraphics[width=3.7cm]{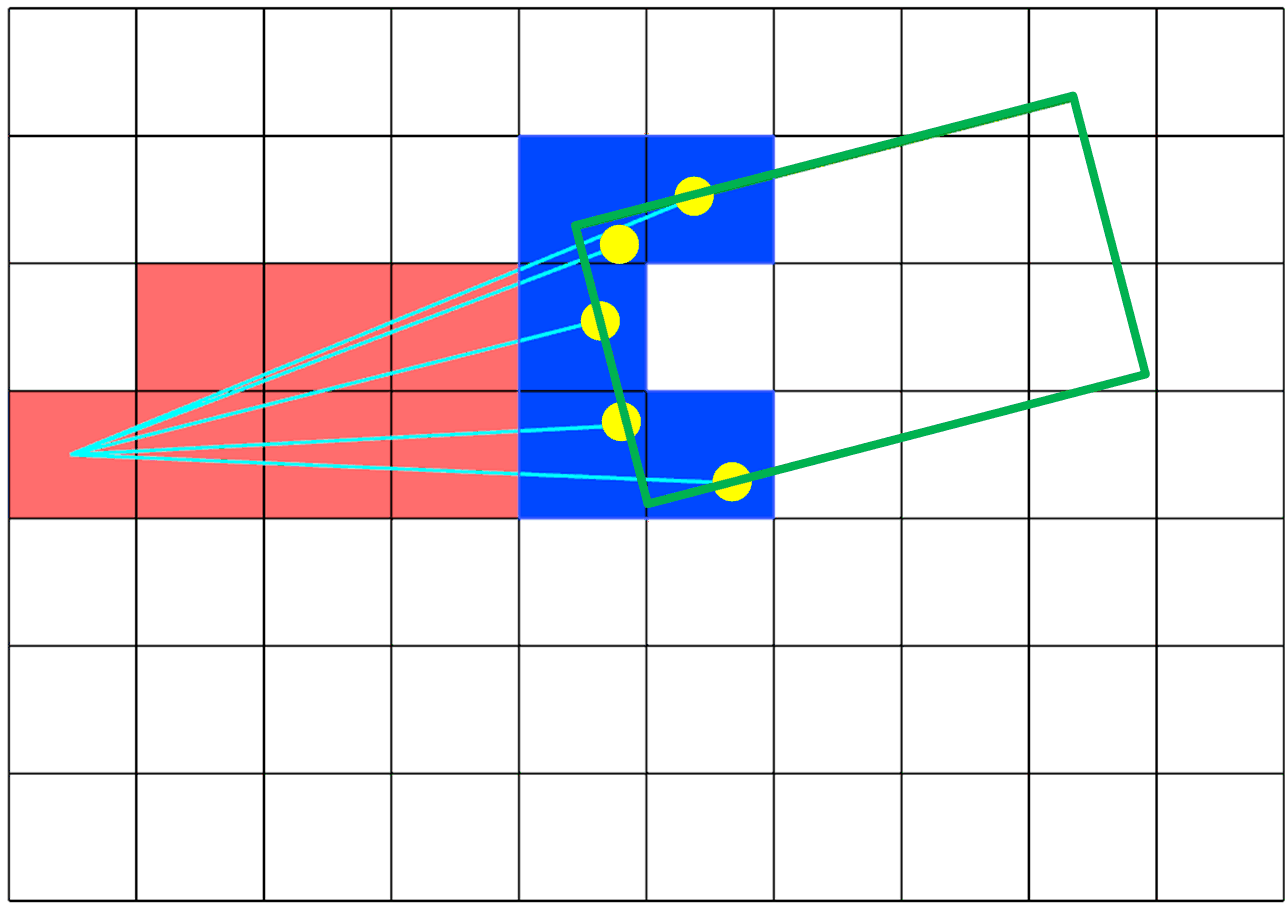}
    \caption{Only voxels passed through by the laser rays must be true negative.}
    \label{fig:occ_tn}
  \end{subfigure}
  \caption{An example of a slice of occupancy map at a certain height. For demonstration purposes, voxel sizes and bounding boxes are adjusted from their real settings.}
  \label{fig:occ}
\end{figure}

\subsection{Handcrafted Gaussian Positional Encoding}
\label{subsec:gaussian_positional_encoding}
Generally in attention mechanisms, positional encodings or linearly transformed positional encodings are widely adopted to provide their cosine similarity as a portion of the attention weight to gather values \cite{dai-etal-2019-transformer, JMLR:v21:20-074}. If distance between two positions is observed with random fluctuating noise, we assume that ideally, the probability density of the observed distances between the two actually same positions must obey univariate normal distribution.

\begin{flushleft}
  \textbf{Definition} 
\end{flushleft}

We define the probability density function:

\begin{equation}
  f(x, \mu) = \frac{1}{\sqrt{\pi} \sigma} e^{-\frac{(x-\mu)^2}{\sigma^2}}
\end{equation}

We define our positional encoding function $g(d)$ which transforms position $d \in [d_{min}, d_{max}]$ to a vector.

\begin{equation}
  g(d) = [f(x_1, d) \sqrt{\Delta x}, ..., f(x_n, d) \sqrt{\Delta x}]
  \label{eq:gaussian_possitional encoding}
\end{equation}

where $\{x_1, ..., x_n \}$ is a set of $n$ uniformly spaced points over the interval $[x_1, x_n]$, $x_1 \ll d_{min}$ and $x_n \gg d_{max}$.

\begin{flushleft}
  \textbf{Claim}
\end{flushleft}

For $\Delta d$ with its probability density function defined as the cosine similarity of $g(d_1)$ and $g(d_2)$ (see \Cref{eq:cosine_sim}), $\Delta d \sim \mathcal{N}(0; \sigma)$.

\begin{equation}
  P(\Delta d) = g(d_1) \cdot g(d_2)
  \label{eq:cosine_sim}
\end{equation}
\begin{flushleft}
  \textbf{Proof}
\end{flushleft}
\begin{dmath}
  P(\Delta d) = g(d_1) \cdot g(d_2) \approx \int ^{+\infty} _{-\infty} f(x, d_1) \cdot f(x, d_2) dx = \frac{1}{\sigma \sqrt{2 \pi}} e^{-\frac{{\Delta d}^2}{2 \sigma^2}}
\end{dmath}
Hence, $\Delta d \sim \mathcal{N}(0; \sigma)$.
\begin{flushleft}
  \textbf{Q.E.D.}
\end{flushleft}

Trough scaling the vector in \Cref{eq:gaussian_possitional encoding} with a constant scaler $(\sqrt{2 \pi} \sigma)^{\frac{1}{2}}$ that normalizes the range of $g(d_1) \cdot g(d_2)$ to $(0, 1]$, we get our positional encoding. In practice, $\sigma$ can be a hyperparameter or a learned parameter of a model.

\subsection{Auxiliary Occupancy Supervision with True Negatives}
The occupancy map is composed of 3D voxels in the form of a 3D matrix where each element indicates whether a voxel is occupied by any objects. Inspired by PLUMENet~\cite{Wang2021PLUMENetE3}, our model estimates the occupancy map as an auxiliary target to avoid overfitting with tighter constraints. To produce ground truth for the occupancy map, one way is to take voxels containing lidar points as positive samples and voxels without points as empty space. However, voxels occupied by an object but not visible to lidar will be false negative samples (see \Cref{fig:occ_fn}). To fix this problem, only voxels passed through by laser beams and without lidar points are chosen as negative samples as \Cref{fig:occ_tn} shows. Though the constraint is looser having fewer negative samples, it can produce labels without false negatives.

\subsection{Multi-task Heads}
For 3D detection head, we deploy settings from the Single Shot Detector(SSD)~\cite{Liu2016SSDSS} and assign ground truth to axis-aligned anchor boxes using 2D Intersection over Union (IoU)~\cite{Everingham2009ThePV} following PointPillars~\cite{Lang2019PointPillarsFE}. Parallel to the detection head, we use the PLUMENet~\cite{Wang2021PLUMENetE3} setup of the occupancy head to predict the occupancy map.

\begin{table*}
  \centering
  \begin{tabular}{l|ccc|ccc}
    \toprule
    \multirow{2}{*}{Method} & \multicolumn{3}{c|}{3D AP@0.7} & \multicolumn{3}{c}{BEV AP@0.7}   \\
                        & Easy & Mod.  & Hard & Easy  & Mod.    & Hard  \\
    \midrule
    OFT-Net~\cite{BMVC2019}        & 2.50      & 3.28      & 2.27      & 9.50      & 7.99       & 7.51      \\
    CaDDN~\cite{reading2021categorical}           & 19.17     & 13.41     & 11.46     & 27.94     & 18.91      & 17.19     \\
    AM3D~\cite{ma2019accurate}           & 21.48     & 16.08     & 15.26     & N/A       & N/A        & N/A       \\
    MonoDTR~\cite{Huang2022MonoDTRM3}       & 21.99     & 15.39     & 12.73     & 28.59     & 20.38      & 17.14     \\
    MonoDETR~\cite{Zhang2022MonoDETRDT}      & 24.52     & 16.26     & 13.93     & 32.20     & 21.45      & 18.68     \\
    MonoCon~\cite{Liu2021LearningAM}       & 22.50     & 16.46     & 13.95     & 31.12     & 22.10      & 19.00     \\
    DD3D~\cite{Park2021IsPN}           & 23.19     & 16.87     & 14.36     & 32.35     & 23.41      & 20.42     \\
    MonoDDE~\cite{Li2022DiversityMF}       & 24.93     & 17.14     & 15.10     & 33.58     & 23.46      & 20.37     \\
    GCDR+DDA~\cite{Hu2022DeepLM}       & 25.21     & 17.25     & 13.53     & \emph{37.86}   & \emph{27.92}    & \emph{21.97}   \\
    PS-fld~\cite{Chen2022PseudoStereoFM}        & 23.74     & 17.74     & 15.14     & 32.64     & 23.76      & 20.64     \\
    LPCG-Monoflex~\cite{Peng2021LidarPC} & 25.56     & 17.80     & 15.38     & 35.96     & 24.81      & 21.86     \\
    MonoInsight~\cite{Haq2022OneSM}   & \emph{27.71}   & \emph{19.04}   & \emph{16.03}   & 34.85     & 24.23      & 20.87     \\
    Ours               & \bf{31.55} & \bf{20.95} & \bf{17.83} & \bf{41.41} & \bf{28.50}  & \bf{23.88} \\
    \hline
    LPCG + DM3D~\cite{peng2022digging}     & \emph{29.15}   & \emph{21.24}   & \emph{19.18}   & \emph{39.74}   & \emph{28.84}    & \emph{26.08}   \\
    Ours + DM3D        & \bf{35.96} & \bf{25.02} & \bf{21.47} & \bf{46.17} & \bf{33.13}  & \bf{28.80} \\
    \bottomrule
  \end{tabular}
  \caption{Object detection results for car category ($AP|_{R_{40}}$ metric~\cite{Simonelli_2019_ICCV}) on the KITTI~\cite{Geiger2012AreWR} test set. Our model outperforms all previously published state-of-the-art one-stage and Pseudo-LiDAR methods in both 3D and BEV benchmarks for all difficulty levels. The best results are marked in bold and the second-best results are in italics. DM3D~\cite{peng2022digging} though not practical for autonomous robots by just producing more boxes to increase recall, improves AP significantly. For a fair comparison, we also report DM3D applied to our model verse the best model in its original paper, LPCG + DM3D~\cite{peng2022digging}. Our results are still the best for all metrics.}
  \label{tab:benchmark_evaluation}
\end{table*}

\section{Experiments}
\label{sec:Experiments}
\newcolumntype{P}{>{\raggedleft\arraybackslash}p{0.5cm}}

\subsection{KITTI Object Detection Dataset}
KITTI is a pioneer dataset for visual recognition systems in robotic applications~\cite{Geiger2012AreWR}. The data is captured from a VW station wagon equipped with four video cameras (1392$\times$512 pixels), a Velodyne 64-beam laser scanner, and a localization unit. KITTI provides 7481 training samples and 7518 testing samples, each containing two color images, two grayscale images, a point cloud, object annotations in the form of 3D bounding boxes, and a calibration file. Only annotations of the training set are made public. A total of 80256 labeled objects are divided into 7 classes: 'Car', 'Van', 'Truck', 'Pedestrian', 'Person (sitting)', 'Cyclist', 'Tram' and 'Misc'. There are three difficulty levels, easy, moderate, and hard, defined by bounding box height and occlusion level. The 3D object detection benchmark ranks all methods using the average precision of precision-recall curves with 40 recall positions.

\subsection{Implementation Details}
The 2D backbone is composed of a V2-99 net~\cite{Lee2020CenterMaskRA} extended to an FPN and two branches to generate image values and depth keys. The V2-99 net is pretrained on DDAD15M~\cite{Guizilini20203DPF} as DD3D~\cite{Park2021IsPN} for depth estimation.  Multi-scale features from the FPN are bilinearly interpolated to the same size  and concatenated as input to the two branches. The BEV backbone is a 3D-BEV network in PLUMENet-Small~\cite{Wang2021PLUMENetE3}.

All input images are resized to $352 \times 1248$. Each image value and depth key has a size of $64 \times 88 \times 312$. For the Gaussian encoding function generating the queries, we set $n=64$, $x_1=0$, $x_n=64.8$. For another Gaussian encoding function used to encode the orientation difference, we set $n=64$, $x_1=-0.9$, $x_n=0.9$. For benchmarking and ablation studies, voxels are $0.4 \times 0.4 \times 0.4$. We use voxel sizes of $0.2 \times 0.2 \times 0.2$ in a tiny model version.

We trained on the entire training set for evaluation on the KITTI test server. While for the ablation study, we trained on the subset of the training set the same as SECOND~\cite{yan2018second}. All models are trained 80 epochs using AdamW~\cite{Loshchilov2019DecoupledWD} optimizer with a one-cycle learning rate scheduler~\cite{Smith2019SuperconvergenceVF}. The maximum learning rate is set to 0.0003, and the dividing factor is 20.

For data augmentation, we randomly select an image captured by a left or right color camera for each sample; we deploy random world flipping and random world rotation with range ($-\frac{\pi}{4}, \frac{\pi}{4}$) as described in \cref{subsec:orientation_aware}. The lidar coordinate system in KITTI is chosen as the ego-car coordinate system.

\subsection{Benchmark Evaluation}
\noindent \fontsize{10pt}{\baselineskip}\selectfont {\bf Comparison with State-of-the-Art Methods.}
To make a horizontal comparison of monocular models, we evaluate our result of the test set on the KITTI server. The results are summarized in \Cref{tab:benchmark_evaluation}. On the KITTI 3D object detection benchmark, our model ranks first in all difficulty levels among the previous monocular models by achieving 20.95\% AP on Car Moderate. Our model outperforms the previously best 3D intermediate representation method~\cite{Hu2022DeepLM} by a large margin of 3.15\%. On the KITTI bird's eye view benchmark, our model achieves an AP Moderate score of 28.50\%, outperforming all published models.
\setlength{\parskip}{5pt}

\noindent \fontsize{10pt}{\baselineskip}\selectfont {\bf Results on Pedestrian.}
As shown in \Cref{tab:pedestrian_results} We evaluate $AP_{BEV}$ and $AP_{3D}$ results on both KITTI test set and KITTI val set as some prior works did not report pedestrian results to KITTI test server. We ranks the 1st among all mothods using an intermediate 3D representation. Despite that pedestrian is smaller in bird's eye view than from a frontal view which makes detection more challenging, our method outperforms most one-stage frontal-view methods and achieves competitive results to DD3D~\cite{Park2021IsPN}.
\setlength{\parskip}{0pt}

\begin{table}
  \centering
  \begin{tabular}{p{1.9cm}|PP>{\raggedleft\arraybackslash}p{0.75cm} | PP>{\raggedleft\arraybackslash}p{0.75cm}}
    \toprule
    \multirow{2}{*}{Method} & \multicolumn{3}{c|}{3D AP@0.7} & \multicolumn{3}{c}{3D AP@0.5} \\    
          & Easy     & Mod.     & Hard     & Easy     & Mod.     & Hard  \\
    \midrule
    Sinusoidal* & 25.24 &	19.07 &	\emph{16.38} &	65.28 &	\bf{53.85} &	\emph{47.07} \\
    Learned*    & \emph{26.30} &	\emph{19.46} &	16.37 &	\bf{69.09} &	\emph{53.29} &	46.79 \\
    Gaussian*   & 24.47 &	17.55 &	14.83 &	67.25 &	50.72 &	46.04 \\
    Sinusoidal & 24.74 & 18.79 & 16.06 & 68.53 & 52.82 & 46.86 \\
    Learned & N/A & N/A & N/A & N/A & N/A & N/A \\
    Gaussian & \bf{27.05} & \bf{19.72} & \bf{17.16} & \emph{68.53} & 53.14 & \bf{47.09} \\
    \bottomrule
  \end{tabular}
  \caption{Ablation experiments for positional encodings. On par with the learned method in performance, our encoding function has the advantage of encoding continuous values. Our Gaussian positional encoding is superior to the sinusoidal method to a large extent for AP@0.7. *Sigmoid activation is applied.}
  \label{tab:positional_encoding}
\end{table}
\begin{table}
  \centering
  \begin{tabular}{p{1.9cm}|PP>{\raggedleft\arraybackslash}p{0.75cm} | PP>{\raggedleft\arraybackslash}p{0.75cm}}
    \toprule
    \multirow{2}{*}{Method} & \multicolumn{3}{c|}{3D AP@0.7} & \multicolumn{3}{c}{3D AP@0.5} \\ 
    & Easy     & Mod. & Hard  & Easy     & Mod. & Hard  \\
    \midrule
    Baseline & 23.40 & 15.55 & 12.85 & 65.02 & 44.69 & 36.64 \\
    +AUG     & 23.12 & 15.23 & 12.55 & 65.59 & 45.15 & 38.25 \\
    +OAB     & 27.05 & 19.72 & 17.16 & 68.53 & 53.14 & 47.09 \\
    +AUG+OAB & \bf{38.25} & \bf{27.60} & \bf{23.48} & \bf{78.95} & \bf{63.16} & \bf{56.09} \\
    \bottomrule
  \end{tabular}
  \caption{Ablation experiments for the orientation-aware backbone. "AUG" indicates the random world rotation and flipping augmentation. "OAB" represents the orientation-aware backbone. OAB is a necessity for AUG to boost model performance.}
  \label{tab:orientation_aware}
\end{table}
\begin{table}
  \centering
  \resizebox{\linewidth}{!}{
    \begin{tabular}{p{1.9cm}|p{0.6cm}p{0.6cm}|PP>{\raggedleft\arraybackslash}p{0.75cm} | PP>{\raggedleft\arraybackslash}p{0.75cm}}
      \toprule
      \multirow{2}{*}{Method} & \multirow{2}{*}{AUG} & \multirow{2}{*}{OAB} & \multicolumn{3}{c|}{3D AP@0.7} & \multicolumn{3}{c}{3D AP@0.5} \\ 
      & & & Easy     & Mod. & Hard  & Easy     & Mod. & Hard  \\
      \midrule
      Projection & & & 5.35   & 3.49  & 2.8   & 35.45    & 23.68    & 19.94 \\
      LRA & & & 23.40 & 15.55 & 12.85 & 65.02 & 44.69 & 36.64 \\
      Projection & \checkmark & & 26.57 & 19.68 & 16.75 & 66.42 & 50.89 & 44.84 \\
      LSS~\cite{10.1007/978-3-030-58568-6_12} & \checkmark & & 19.89 & 15.30 & 13.53 & 64.19 & 49.68 & 44.24 \\
      LRA & \checkmark & & 27.05 & 19.72 & 17.15 & 68.53 & 53.14 & 47.09 \\
      Projection & \checkmark & \checkmark & 37.09 & 26.96 & 22.94 & 75.44 & 60.17 & 53.17 \\
      LSS & \checkmark & \checkmark & 31.95 & 22.90 & 19.68 & 73.36 & 59.30 & 53.20 \\
      LRA & \checkmark & \checkmark & \bf{38.25} & \bf{27.60} & \bf{23.48} & \bf{78.95} & \bf{63.16} & \bf{56.09} \\
      \bottomrule
    \end{tabular}
  }
  \caption{Ablation experiments for choices of transforming 2D features to 3D space. "LRA" indicates the local ray attention.}
  \label{tab:local_ray_attention}
\end{table}
\begin{table}
  \centering
  \resizebox{\linewidth}{!}{
    \begin{tabular}{ccccc}
      \toprule
      Method & \textbf{DS} & Easy & Moderate & Hard \\
      \midrule
      PLiDAR~\cite{weng2019monocular} & Val & 14.4 / 11.6 & 13.8 / 11.2 & 12.0 / 10.9 \\
      Ours & Val & \bf{20.68} / \bf{17.90} & \bf{15.17} / \bf{12.87} & \bf{12.41} / \bf{10.59} \\
      \midrule
      DD3D~\cite{Park2021IsPN} & Test & \bf{17.74} / \bf{16.64} & \bf{12.16} / \bf{11.04} & \bf{10.49} / \bf{9.38} \\
      Ps-fld~\cite{Chen2022PseudoStereoFM} & Test & 12.80/ 11.22 & 7.29 / 6.18 & 6.05 / 5.21 \\
      MonoDDE~\cite{Li2022DiversityMF} & Test & 12.38 / 11.13 & 8.41 / 7.32 & 7.16 / 6.67 \\
      \midrule
      OFTNet~\cite{BMVC2019} & Test & 1.28 / 0.63 & 0.81 / 0.36 & 0.51 / 0.35 \\
      CaDDN~\cite{reading2021categorical} & Test & 14.72 / 12.87 & 9.41 / 8.14 & 8.17 / 6.76 \\
      Ours & Test & \bf{17.90}  / \bf{16.19} & \bf{11.94} / \bf{10.53} & \bf{10.34} / \bf{8.97} \\
      \bottomrule
    \end{tabular}
  }
  \caption{$AP_{BEV}$ / $AP_{3D}$ results for pedestrians at IoU = 0.5.}
  \label{tab:pedestrian_results}
\end{table}
\begin{table}
  \centering
  \begin{tabular}{ccccc}
    \toprule
    Voxel Size & Easy & Mod. & Hard & Inference Time \\
    \midrule
    0.2m & 38.25 & 27.60 & 23.48 & 129.39ms \\
    0.4m & 33.50 & 24.07 & 20.60 & 39.49ms \\
    \bottomrule
  \end{tabular}
  \caption{$AP_{3D}$ scores at IoU=0.7 varying the voxel size on the KITTI validation set.}
  \label{tab:inference time}
\end{table}

\subsection{Ablation Study}
To demonstrate the importance of each component of our model, we measure the change in average precision evaluated on the KITTI validation set while the model is modified.

\begin{table*}
  \centering
  \begin{tabular}{c|ccc|ccc|ccc|ccc}
    \toprule
    \multirow{2}{*}{Method} & \multicolumn{3}{c|}{3D AP@0.7} & \multicolumn{3}{c|}{BEV AP@0.7} & \multicolumn{3}{c|}{3D AP@0.5} & \multicolumn{3}{|c}{BEV AP@0.5} \\ 
    & Easy     & Mod. & Hard  & Easy     & Mod. & Hard  & Easy     & Mod. & Hard  & Easy     & Mod. & Hard\\
    \midrule
    -{\bf OCC.} Map with {\bf TN}      & 34.59 & 24.74 & 21.08 &	45.13 & 32.90 &	28.50 &	74.96 &	58.92 &	52.03 &	78.79 &	64.03 &	56.95 \\
    -Local Ray \bf{Att}.     & 37.09 &	26.96 &	22.94 &	47.81 &	34.92 &	30.38 &	75.44 &	60.17 &	53.17 &	78.78 &	65.29 &	58.99 \\
    -Orientation-aware {\bf BB.}  & 23.12 &	15.23 &	12.55 &	34.44 &	23.35 &	19.39 &	65.59 &	45.15 &	38.25 &	69.88 &	48.42 &	41.96 \\
    Full Model          & \bf{38.25}	&	\bf{27.60} & \bf{23.48}	&	\bf{49.58}	&	\bf{36.05}	&	\bf{31.26}	&	\bf{78.95}	&	\bf{63.16}	&	\bf{56.09}	&	\bf{81.76}	&	\bf{66.42}	&	\bf{60.43} \\
    \bottomrule
  \end{tabular}
  \caption{Summarization of ablation studies for key contributions with more detailed metrics. Removing any of these components impairs the performance of the model.}
  \label{tab:key_contributions}
\end{table*}
\begin{figure*}
  \centering
  \includegraphics[width=\textwidth]{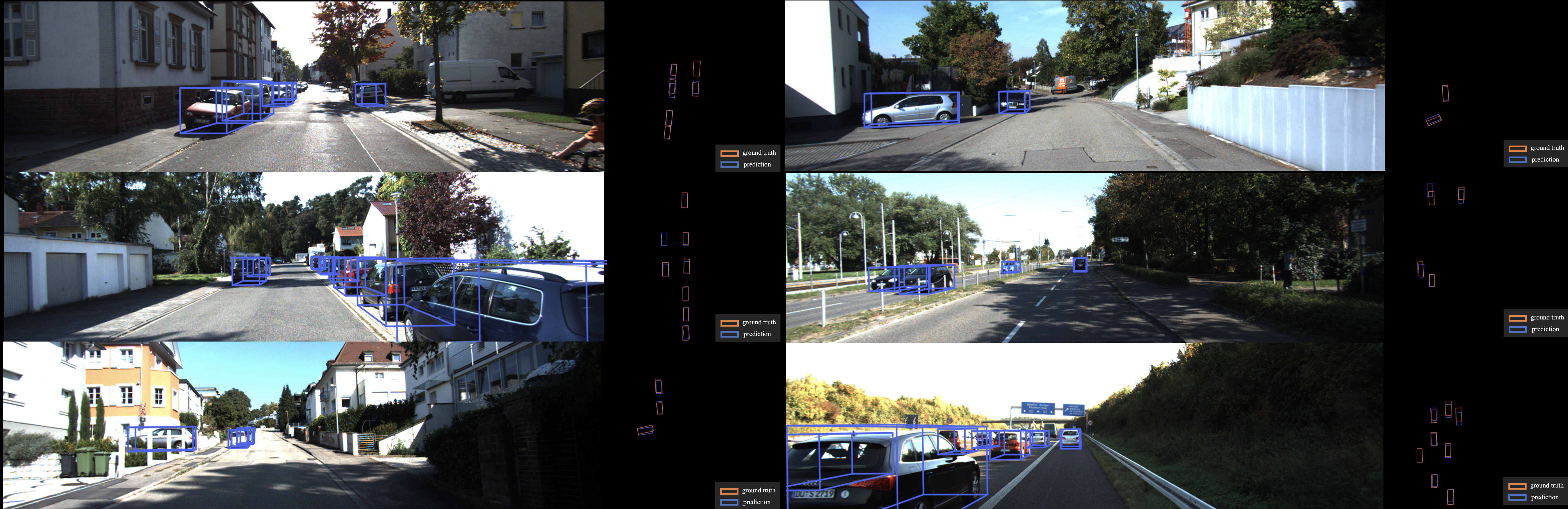}
  \caption{Qualitative examples from the validation set in KITTI. The blue bounding box represents the predicted bounding box for each car. Bird's-eye view is also provided to show that our model can accurately recover the distance of objects. Note that all these images are not included in the training phase.}
  \label{fig:qualitative_results}
\end{figure*}

\noindent \fontsize{10pt}{\baselineskip}\selectfont {\bf Positional Encodings.}
In \Cref{tab:positional_encoding}, we replace Gaussian positional encodings with sinusoidal~\cite{NIPS2017_3f5ee243} and learned positional encodings. We do not apply random world rotation to any of our models because this would greatly vary the distance from each voxel to the image plane, which would require a continuous positional encoding function instead of a discrete learned function. Since the learned way can not guarantee the dot product of arbitrary positional encodings to have a proper upper bound not too large, the version without sigmoid activation diverged during training. Gaussian positional encoding without sigmoid activation surpasses the other methods in most metrics. While Gaussian and learned ways obtain similar performance for AP@0.5 metrics, the Gaussian encoding prevails over learned encoding in its ability to represent continuous value rather than a fixed amount of discrete values.
\setlength{\parskip}{5pt}

\noindent \fontsize{10pt}{\baselineskip}\selectfont {\bf Local Ray Attention Mechanism.}
To verify the effectiveness of the local ray attention mechanism, we vary the 2D-3D transformation components as in \Cref{tab:local_ray_attention}. Interestingly, we observe that removing local ray attention ("Projection" in \Cref{fig:local_ray_attention}: Directly project voxel centers to the image plane to fetch features) hurts model quality considerably without an orientation-aware backbone. Applying an orientation-aware backbone dramatically reduces the gap between configurations with and without a local ray attention mechanism. Additionally, LSS~\cite{10.1007/978-3-030-58568-6_12} is not superior to voxel projection, perhaps due to the sparsity of distant pseudo-points.

\noindent \fontsize{10pt}{\baselineskip}\selectfont {\bf Orientation-Aware 2D Image Backbone.}
In \Cref{tab:orientation_aware}, we observe that applying the random world rotation and flipping augmentation hardly brings any gain of AP. By deploying an orientation-aware backbone alone, we obtain at least 3.51\% increase for all metrics even if augmentation is not applied. The augmentation further improves the 3D moderate result remarkably to the extent of 12.05\% with the assistance of the orientation-aware backbone. This suggests that ensuring implicit and explicit features are consistent is vital in transforming 2D features into 3D space.

\noindent \fontsize{10pt}{\baselineskip}\selectfont {\bf Occupancy Supervision with True Negative.}
Additionally, we experimented with the two label assignment methods of the occupancy map. The baseline uses the label assignment scheme as \Cref{fig:occ_fn}. As expected, though the occupancy map with true negative annotates fewer voxels resulting in looser supervision, it is very helpful in avoiding over-fitting by providing labels more correctly (see \Cref{tab:key_contributions}).
\setlength{\parskip}{0pt}

\subsection{Inference Time}
We present two versions of the models in \Cref{tab:inference time} that differ in the voxel size of intermediate representations. 1. The tiny version can be processed in approximately 25.3 FPS, which is sufficient for most real-time applications. With a fine-grained voxel partition, the base version requires 129.39ms to infer, which has a higher average precision outperforming all reported methods.

\subsection{Qualitative Results}
The qualitative results on \textbf{val} set are shown in \Cref{fig:qualitative_results}. For better visualization and comparison, we plot the ground-truth bounding box and predicted bounding box in a bird's eye view. It is clear from our results that even when cars are largely occluded in the image, our model is able to detect them accurately.

\section{Conclusion}
\label{sec:Conclusion}
In this work, we identify the inconsistency of implicit and explicit features as the primary obstacle that hinders the development of methods using intermediate 3D representation. Based on this insight, we propose a novel detection paradigm with the orientation-aware image backbone, the local ray attention mechanism, and the handcrafted Gaussian encoding. Extensive ablation experiments show that ensuring the consistency improves the performance of our model significantly. As expected, the local ray attention mechanism and Gaussian encoding have proven to be effective. Overall, our model put a spurt in detection accuracy in KITTI 3D and BEV benchmarks showing promising results.

{\small
\bibliographystyle{ieee_fullname}
\bibliography{egbib}

\begin{thebibliography}{10}\itemsep=-1pt

\bibitem{carion2020end}
Nicolas Carion, Francisco Massa, Gabriel Synnaeve, Nicolas Usunier, Alexander
  Kirillov, and Sergey Zagoruyko.
\newblock End-to-end object detection with transformers.
\newblock In {\em European conference on computer vision}, pages 213--229.
  Springer, 2020.

\bibitem{monopair}
Yongjian Chen, Lei Tai, Kai Sun, and Mingyang Li.
\newblock Monopair: Monocular 3d object detection using pairwise spatial
  relationships.
\newblock In {\em 2020 IEEE/CVF Conference on Computer Vision and Pattern
  Recognition (CVPR)}, pages 12090--12099, 2020.

\bibitem{dai-etal-2019-transformer}
Zihang Dai, Zhilin Yang, Yiming Yang, Jaime Carbonell, Quoc Le, and Ruslan
  Salakhutdinov.
\newblock Transformer-{XL}: Attentive language models beyond a fixed-length
  context.
\newblock In {\em Proceedings of the 57th Annual Meeting of the Association for
  Computational Linguistics}, pages 2978--2988, Florence, Italy, July 2019.
  Association for Computational Linguistics.

\bibitem{Everingham2009ThePV}
Mark Everingham, Luc~Van Gool, Christopher K.~I. Williams, John~M. Winn, and
  Andrew Zisserman.
\newblock The pascal visual object classes (voc) challenge.
\newblock {\em International Journal of Computer Vision}, 88:303--338, 2009.

\bibitem{Geiger2012AreWR}
Andreas Geiger, Philip Lenz, and Raquel Urtasun.
\newblock Are we ready for autonomous driving? the kitti vision benchmark
  suite.
\newblock {\em 2012 IEEE Conference on Computer Vision and Pattern
  Recognition}, pages 3354--3361, 2012.

\bibitem{Guizilini20203DPF}
Vitor~Campanholo Guizilini, Rares Ambrus, Sudeep Pillai, and Adrien Gaidon.
\newblock 3d packing for self-supervised monocular depth estimation.
\newblock {\em 2020 IEEE/CVF Conference on Computer Vision and Pattern
  Recognition (CVPR)}, pages 2482--2491, 2020.

\bibitem{Haq2022OneSM}
Muhamad~Amirul Haq, Shanq-Jang Ruan, Mei-En Shao, Qazi~Mazhar ul Haq, Pei-Jung
  Liang, and De-Qin Gao.
\newblock One stage monocular 3d object detection utilizing discrete depth and
  orientation representation.
\newblock {\em IEEE Transactions on Intelligent Transportation Systems}, 2022.

\bibitem{Hu2022DeepLM}
Henan Hu, Ming Zhu, Muyu Li, and Kwok-Leung Chan.
\newblock Deep learning-based monocular 3d object detection with refinement of
  depth information.
\newblock {\em Sensors (Basel, Switzerland)}, 22, 2022.

\bibitem{Huang2021BEVDetHM}
Junjie Huang, Guan Huang, Zheng Zhu, and Dalong Du.
\newblock Bevdet: High-performance multi-camera 3d object detection in
  bird-eye-view.
\newblock {\em ArXiv}, abs/2112.11790, 2021.

\bibitem{Huang2022MonoDTRM3}
Kuan-Chih Huang, Tsung-Han Wu, Hung-Ting Su, and Winston~H. Hsu.
\newblock Monodtr: Monocular 3d object detection with depth-aware transformer.
\newblock {\em ArXiv}, abs/2203.10981, 2022.

\bibitem{Lang2019PointPillarsFE}
Alex~H. Lang, Sourabh Vora, Holger Caesar, Lubing Zhou, Jiong Yang, and Oscar
  Beijbom.
\newblock Pointpillars: Fast encoders for object detection from point clouds.
\newblock {\em 2019 IEEE/CVF Conference on Computer Vision and Pattern
  Recognition (CVPR)}, pages 12689--12697, 2019.

\bibitem{Lee2020CenterMaskRA}
Youngwan Lee and Jongyoul Park.
\newblock Centermask: Real-time anchor-free instance segmentation.
\newblock {\em 2020 IEEE/CVF Conference on Computer Vision and Pattern
  Recognition (CVPR)}, pages 13903--13912, 2020.

\bibitem{Li2022DiversityMF}
Zhuoling Li, Z. Qu, Yang Zhou, Jianzhuang Liu, Haoqian Wang, and Lihui Jiang.
\newblock Diversity matters: Fully exploiting depth clues for reliable
  monocular 3d object detection.
\newblock {\em ArXiv}, abs/2205.09373, 2022.

\bibitem{li2022bevformer}
Zhiqi Li, Wenhai Wang, Hongyang Li, Enze Xie, Chonghao Sima, Tong Lu, Yu Qiao,
  and Jifeng Dai.
\newblock Bevformer: Learning bird’s-eye-view representation from
  multi-camera images via spatiotemporal transformers.
\newblock {\em arXiv preprint arXiv:2203.17270}, 2022.

\bibitem{Liu2016SSDSS}
W. Liu, Dragomir Anguelov, D. Erhan, Christian Szegedy, Scott~E. Reed,
  Cheng-Yang Fu, and Alexander~C. Berg.
\newblock Ssd: Single shot multibox detector.
\newblock In {\em ECCV}, 2016.

\bibitem{Liu2021LearningAM}
Xianpeng Liu, Nan Xue, and Tianfu Wu.
\newblock Learning auxiliary monocular contexts helps monocular 3d object
  detection.
\newblock {\em ArXiv}, abs/2112.04628, 2021.

\bibitem{liu2020SMOKE}
Zechen Liu, Zizhang Wu, and Roland T\'oth.
\newblock {SMOKE}: Single-stage monocular 3d object detection via keypoint
  estimation.
\newblock {\em arXiv preprint arXiv:2002.10111}, 2020.

\bibitem{Loshchilov2019DecoupledWD}
Ilya Loshchilov and Frank Hutter.
\newblock Decoupled weight decay regularization.
\newblock In {\em ICLR}, 2019.

\bibitem{ma2019accurate}
Xinzhu Ma, Zhihui Wang, Haojie Li, Pengbo Zhang, Wanli Ouyang, and Xin Fan.
\newblock Accurate monocular 3d object detection via color-embedded 3d
  reconstruction for autonomous driving.
\newblock In {\em Proceedings of the IEEE/CVF International Conference on
  Computer Vision}, pages 6851--6860, 2019.

\bibitem{Chen2022PseudoStereoFM}
Yi nan Chen, Hang Dai, and Yong Ding.
\newblock Pseudo-stereo for monocular 3d object detection in autonomous
  driving.
\newblock {\em ArXiv}, abs/2203.02112, 2022.

\bibitem{Park2021IsPN}
Dennis Park, Rares Ambrus, Vitor~Campanholo Guizilini, Jie Li, and Adrien
  Gaidon.
\newblock Is pseudo-lidar needed for monocular 3d object detection?
\newblock {\em 2021 IEEE/CVF International Conference on Computer Vision
  (ICCV)}, pages 3122--3132, 2021.

\bibitem{Peng2021LidarPC}
Liang Peng, Fei Liu, Zhengxu Yu, Senbo Yan, Dan Deng, and Deng Cai.
\newblock Lidar point cloud guided monocular 3d object detection.
\newblock {\em ArXiv}, abs/2104.09035, 2021.

\bibitem{peng2022digging}
Liang Peng, Senbo Yan, Chenxi Huang, Xiaofei He, and Deng Cai.
\newblock Digging into output representation for monocular 3d object detection,
  2022.

\bibitem{10.1007/978-3-030-58568-6_12}
Jonah Philion and Sanja Fidler.
\newblock Lift, splat, shoot: Encoding images from arbitrary camera rigs by
  implicitly unprojecting to 3d.
\newblock In Andrea Vedaldi, Horst Bischof, Thomas Brox, and Jan-Michael Frahm,
  editors, {\em Computer Vision -- ECCV 2020}, pages 194--210, Cham, 2020.
  Springer International Publishing.

\bibitem{qi2017pointnet}
Charles~R Qi, Hao Su, Kaichun Mo, and Leonidas~J Guibas.
\newblock Pointnet: Deep learning on point sets for 3d classification and
  segmentation.
\newblock In {\em Proceedings of the IEEE conference on computer vision and
  pattern recognition}, pages 652--660, 2017.

\bibitem{JMLR:v21:20-074}
Colin Raffel, Noam Shazeer, Adam Roberts, Katherine Lee, Sharan Narang, Michael
  Matena, Yanqi Zhou, Wei Li, and Peter~J. Liu.
\newblock Exploring the limits of transfer learning with a unified text-to-text
  transformer.
\newblock {\em Journal of Machine Learning Research}, 21(140):1--67, 2020.

\bibitem{reading2021categorical}
Cody Reading, Ali Harakeh, Julia Chae, and Steven~L Waslander.
\newblock Categorical depth distribution network for monocular 3d object
  detection.
\newblock In {\em Proceedings of the IEEE/CVF Conference on Computer Vision and
  Pattern Recognition}, pages 8555--8564, 2021.

\bibitem{BMVC2019}
Thomas Roddick, Alex Kendall, and Roberto Cipolla.
\newblock Orthographic feature transform for monocular 3d object detection.
\newblock In Kirill Sidorov and Yulia Hicks, editors, {\em Proceedings of the
  British Machine Vision Conference (BMVC)}, pages 59.1--59.13. BMVA Press,
  September 2019.

\bibitem{Simonelli_2019_ICCV}
Andrea Simonelli, Samuel~Rota Bulo, Lorenzo Porzi, Manuel Lopez-Antequera, and
  Peter Kontschieder.
\newblock Disentangling monocular 3d object detection.
\newblock In {\em Proceedings of the IEEE/CVF International Conference on
  Computer Vision (ICCV)}, October 2019.

\bibitem{Smith2019SuperconvergenceVF}
Leslie~N. Smith and Nicholay Topin.
\newblock Super-convergence: very fast training of neural networks using large
  learning rates.
\newblock In {\em Defense + Commercial Sensing}, 2019.

\bibitem{tian2019fcos}
Zhi Tian, Chunhua Shen, Hao Chen, and Tong He.
\newblock {FCOS}: Fully convolutional one-stage object detection.
\newblock In {\em Proc. Int. Conf. Computer Vision (ICCV)}, 2019.

\bibitem{NIPS2017_3f5ee243}
Ashish Vaswani, Noam Shazeer, Niki Parmar, Jakob Uszkoreit, Llion Jones,
  Aidan~N Gomez, \L~ukasz Kaiser, and Illia Polosukhin.
\newblock Attention is all you need.
\newblock In I. Guyon, U.~Von Luxburg, S. Bengio, H. Wallach, R. Fergus, S.
  Vishwanathan, and R. Garnett, editors, {\em Advances in Neural Information
  Processing Systems}, volume~30. Curran Associates, Inc., 2017.

\bibitem{wang2021fcos3d}
Tai Wang, Xinge Zhu, Jiangmiao Pang, and Dahua Lin.
\newblock {FCOS3D: Fully} convolutional one-stage monocular 3d object
  detection.
\newblock In {\em Proceedings of the IEEE/CVF International Conference on
  Computer Vision (ICCV) Workshops}, 2021.

\bibitem{detr3d}
Yue Wang, Vitor Guizilini, Tianyuan Zhang, Yilun Wang, Hang Zhao, , and
  Justin~M. Solomon.
\newblock Detr3d: 3d object detection from multi-view images via 3d-to-2d
  queries.
\newblock In {\em The Conference on Robot Learning ({CoRL})}, 2021.

\bibitem{Wang2021PLUMENetE3}
Yan Wang, Binh Yang, Rui Hu, Ming Liang, and Raquel Urtasun.
\newblock Plumenet: Efficient 3d object detection from stereo images.
\newblock {\em 2021 IEEE/RSJ International Conference on Intelligent Robots and
  Systems (IROS)}, pages 3383--3390, 2021.

\bibitem{weng2019monocular}
Xinshuo Weng and Kris Kitani.
\newblock Monocular 3d object detection with pseudo-lidar point cloud.
\newblock In {\em Proceedings of the IEEE/CVF International Conference on
  Computer Vision Workshops}, pages 0--0, 2019.

\bibitem{yan2018second}
Yan Yan, Yuxing Mao, and Bo Li.
\newblock Second: Sparsely embedded convolutional detection.
\newblock {\em Sensors}, 18(10):3337, 2018.

\bibitem{Zhang2022MonoDETRDT}
Renrui Zhang, Hang Qiu, Tai Wang, Xuan Xu, Ziyu Guo, Yu~Jiao Qiao, Peng Gao,
  and Hongsheng Li.
\newblock Monodetr: Depth-aware transformer for monocular 3d object detection.
\newblock {\em ArXiv}, abs/2203.13310, 2022.

\bibitem{zhou2019objects}
Xingyi Zhou, Dequan Wang, and Philipp Kr{\"a}henb{\"u}hl.
\newblock Objects as points.
\newblock In {\em arXiv preprint arXiv:1904.07850}, 2019.

\end{thebibliography}
}

\end{document}